\documentclass[11pt, a4paper]{article}

\usepackage{cancel}
\usepackage{url}
\usepackage[margin=2.5cm]{geometry}
\usepackage{booktabs}
\usepackage{graphicx}
\usepackage{YK}
\usepackage{mathrsfs}
\usepackage{xcolor}

\usepackage[utf8]{inputenc} %
\usepackage[T1]{fontenc}    %
\usepackage{hyperref}       %
\usepackage{url}            %
\usepackage{booktabs}       %
\usepackage{amsfonts}       %
\usepackage{nicefrac}       %
\usepackage{microtype}      %
\usepackage{xcolor}         %
\usepackage[utf8]{inputenc} %
\usepackage[T1]{fontenc}    %
\usepackage{hyperref}       %
\usepackage{url}            %
\usepackage{booktabs}       %
\usepackage{amsfonts}       %
\usepackage{nicefrac}       %
\usepackage{microtype}      %
\usepackage{xcolor}         %
\usepackage{algorithm}
\usepackage{algpseudocode}
\usepackage{graphicx}
\usepackage{csquotes}
\usepackage[english]{babel}
\usepackage{amsthm}
\newtheorem{lem}{Lemma}

\newtheorem{assump}{Assumption}
\title{CPSample: Classifier Protected Sampling for \\Guarding Training Data During Diffusion}
\usepackage[]{caption}
\usepackage{subcaption}
\usepackage{amsmath}
\usepackage[flushleft]{threeparttable}
\usepackage{wrapfig}
\usepackage{array} %

\usepackage[backend=biber,style=ieee,citestyle=ieee]{biblatex}
\begin{document}

\title{\textbf{CPSample: Classifier Protected Sampling for \\Guarding Training Data During Diffusion}}
\author{Joshua Kazdan${}^{1}$, Hao Sun${}^{2}$, Jiaqi Han${}^{2}$, Felix Petersen${}^{2}$, Stefano Ermon${}^{2}$ \\[.4em]
$^{1}$Department of Statistics, Stanford University \\
$^2$Department of Computer Science, Stanford University \\
}

\newcommand{\fpc}[1]{{\color{red!70!black}\textbf{FP: #1}}}

\newcommand{\hao}[1]{\textcolor{purple}{[#1]}}
\newcommand{\se}[1]{\textcolor{magenta}{[SE: #1]}}

\maketitle

\begin{abstract}
\noindent
    Diffusion models have a tendency to exactly replicate their training data, especially when trained on small datasets.  Most prior work has sought to mitigate this problem by imposing differential privacy constraints or masking parts of the training data, resulting in a notable substantial decrease in image quality.
    We present CPSample, a method that modifies the sampling process to prevent training data replication while preserving image quality. 
    CPSample utilizes a classifier that is trained to overfit on random binary labels attached to the training data.
    CPSample then uses classifier guidance to steer the generation process away from the set of points that can be classified with high certainty, a set that includes the training data.
    CPSample achieves FID scores of 4.97 and 2.97 on CIFAR-10 and CelebA-64, respectively, without producing exact replicates of the training data.  
    Unlike prior methods intended to guard the training images, CPSample only requires training a classifier rather than retraining a diffusion model, which is computationally cheaper. 
    Moreover, our technique provides diffusion models with greater robustness against membership inference attacks, wherein an adversary attempts to discern which images were in the model's training dataset.
    We show that CPSample behaves like a built-in rejection sampler, and we demonstrate its capabilities to prevent mode collapse in Stable Diffusion.
\end{abstract}

\section{Introduction}\label{intro}

Diffusion models are an emerging method of image generation that has surpassed GANs on many common benchmarks~\cite{dhariwal2021diffusion}, achieving state-of-the-art FID scores on CIFAR-10~\cite{cifar}, CelebA~\cite{celeba}, ImageNet~\cite{imagenet}, and other touchstone datasets.
Although their capabilities are impressive, diffusion models still suffer from the tendency to exactly replicate images found in their training sets~\cite{carlini, jagielski2023measuring, somepalli2022diffusion}.
This problem is especially pronounced when the training set contains duplicates~\cite{laion}.
Given that diffusion models are sometimes trained on sensitive content, such as patient data~\cite{kazerouni2023diffusion, pinaya2022brain} or copyrighted data~\cite{dhariwal2021diffusion}, this behavior is generally unacceptable. 
Indeed, Google, Midjourney, and Stability AI are already facing lawsuits for using copyrighted data to train image generation models~\cite{other_lawsuits, google_lawsuit}, some of which exactly replicate their training data during inference~\cite{plagiarism}.

The strongest formal guarantee against replicating or revealing training data is differential privacy (DP)~\cite{dwork}. 
Unfortunately, differential privacy is at odds with generation quality.  
Although differentially private training has been implemented for GANs (DP-GAN)~\cite{xie2018differentially}, diffusion models (DPDM, DP-Diffusion)~\cite{dockhorn2023differentially, better_dpdm}, and latent diffusion models (DP-LDM)~\cite{latent_dpdm}, it typically results in significant degradation of image quality.  
Moreover, one cannot easily make a pretrained model differentially private, implying that to achieve differential privacy, one must retrain from scratch.  
This makes negotiating the trade-off between privacy and quality challenging, as trying different levels of privacy requires retraining.
Due to the difficulty of achieving differential privacy while simultaneously maintaining quality, some researchers have pursued more attainable model characteristics that have the same flavor as differential privacy.  
A frequent benchmark for privacy is robustness to membership inference attacks~\cite{hu2023membership}, whereby the attacker aims to infer whether a given image was used to train the model.
Although researchers have devised a multitude of loss and likelihood-based membership inference attacks, so far, there are few existing methods that explicitly aim to defend against these attacks besides differential privacy and data augmentation~\cite{mem_inf, pang2023whitebox}.
A second privacy benchmark measures the cosine similarity in a feature space of a generated image to its nearest neighbor in the training data~\cite{daras2023ambient, douze2024faiss}.  
Ambient diffusion~\cite{daras2023ambient} is one method to prevent excessive similarity to the training data without enforcing differential privacy; however, ambient diffusion still has notable negative effects on FID score.

\begin{wrapfigure}[22]{r}{.4\textwidth}
    \centering
    \vskip -0.1in
    \includegraphics[width=.4\textwidth]{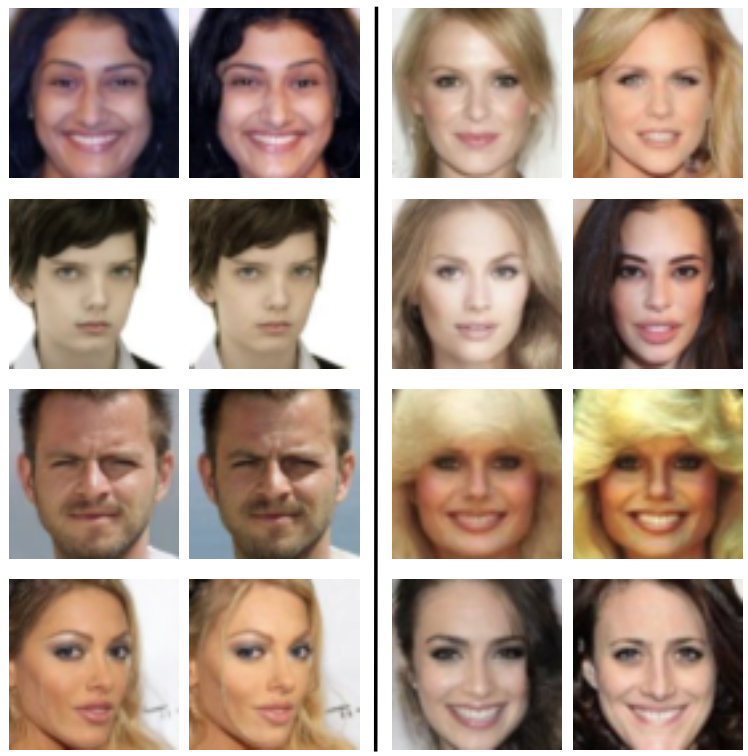}
    \vskip -0.05in
    \caption{
    Generated image and most similar training image pairs for DDIM sampling (left) and CPSample with $\alpha{=}0.001$, $s{=}1000$ (right).  We sample $100$ images and display the four with the highest similarity to their nearest neighbors in the training data.
    }%
    \label{fig:sim_red_celeba}
\end{wrapfigure}

Until recently, preventing image replication by diffusion models has involved various forms of data corruption during training, either by adding noise to gradients~\cite{dpsgd}, diversifying images and captions~\cite{somepalli2023understanding}, or corrupting the images themselves~\cite{daras2023ambient}.  
Hyperparameter tuning for these methods requires retraining, making it difficult to calibrate them to the necessary level of privacy.
Simple alternatives, like rejection sampling, are effective because they can guarantee that the training images will not be exactly replicated.  However, standard rejection sampling has major drawbacks, too.  For instance, rejection sampling redistributes probability mass in an inefficient way and requires resampling, which can decrease speed. In extreme cases of mode collapse such as those uncovered by Webster~\cite{webster2023reproducible}, Stable Diffusion must be queried dozens or even hundreds of times before producing an original image, making rejection sampling impractical. 
Rejection sampling is also prone to membership inference attacks and privacy leakages~\cite{awan2023privacy}.

We present classifier-protected sampling (CPSample), a diffusion-specific data protection technique that, while not strictly differentially private, fortifies against some membership inference attacks and greatly reduces excessive similarity between the training and generated data. 
The basic idea is to overfit a classifier on random binary labels assigned to the training data and use this classifier during sampling to guide the images away from the training data.
We show that our method has an effect similar to rejection sampling while removing or reducing the need to resample.  
Unlike rejection sampling, CPSample offers protection against some membership inference attacks during the generation process rather than only protecting the end product.  %
CPSample achieves SOTA image quality, improving over previous data protection methods,such as ambient diffusion, DPDMs, and PAC Privacy Preserving Diffusion Models~\cite{xu2023pac} for similar levels of ``privacy.'' 
Unlike other methods designed to shield the training data, one can simply adjust the level of protection provided by CPSample without retraining the classifier used for guidance.  
CPSample is applicable to existing image models without any expensive retraining of the diffusion models. 
We summarize the primary contributions of our work as follows:
 \begin{itemize}
     \item In Section~\ref{sampling_method}, we introduce CPSample, a novel method of classifier-guidance for privacy protection in diffusion models that can be applied to existing models without retraining.
     \item We show theoretically in Section~\ref{theory} and empirically in Section~\ref{similarity_reduction_subset} that CPSample prevents training data replication in unguided diffusion.  
     We also provide evidence in Section~\ref{stable_diffusion} that CPSample can protect text-based image generation models like Stable Diffusion.  
     \item We give empirical evidence that CPSample can foil some membership inference attacks in Section~\ref{white_box_inference_attacks}.
     \item We demonstrate in Section~\ref{fid} that CPSample attains better FID scores than existing methods of privacy protection while still eliminating replication of the training data.
 \end{itemize}

\section{Background and Related Work}\label{background}

\subsection{Diffusion Models}
We begin with a review of diffusion models. 
Denoising diffusion probabilistic models (DDPMs)~\cite{sohldickstein2015deep, ho2020denoising} gradually add Gaussian noise to image data during the ``forward'' process.
Meanwhile, one trains a ``denoiser'' to predict the original image from the corrupted samples in a so-called ``backward'' process.
During the forward process, one assigns \begin{equation}
    x_t = \sqrt{\alpha_t}x_0 + \sqrt{1-\alpha_t}\epsilon
\end{equation} where $\epsilon \sim \mathcal{N}(0, I)$, $x_0$ is the original image, and $\alpha_t$ indicates the noise schedule.  The variable $t \in \{0,...,T\}$ specifies the step of the forward process, where $x_0$ represents an image in the training data.  
When $\alpha_T$ is set sufficiently close to $0$, $x_T$ is approximately drawn from a standard normal distribution.
During intermediate steps, the distribution of $x_t$ is \begin{equation}
    q(x_t\mid x_0) = \mathcal{N}(x_t; \sqrt{\alpha_t}x_0, (1-\alpha_t)I).
\end{equation}

During training, one performs gradient descent on $\theta$ to minimize the score-matching loss, given by \begin{equation}
    \mathbb{E}_{\epsilon \sim \mathcal{N}(0,1), x_0 \sim \mathcal{D}}\left[\sum_{t=1}^T\frac{1}{2\sigma_t^2}\|\epsilon - \epsilon_\theta(\sqrt{\overline{\alpha}_t}x_0 + \sqrt{1-\overline{\alpha}_t}\epsilon, t)\|^2\right].
\end{equation}  
Here, $\mathcal{D}$ is the target distribution, which is approximated by sampling from the training data. 
Finally, to generate a new image, one samples standard Gaussian noise $x_T \sim \mathcal{N}(0, I)$.  Then, one gradually denoises $x_T$ by letting \begin{equation}
    x_{t-1} = \frac{1}{\sqrt{\alpha_t}}\left(x_t - \frac{1-\alpha_t}{\sqrt{1-\overline{\alpha}_t}}\epsilon_\theta(x_t, t)\right) + \sigma_t z_t
\end{equation} In each step, $z_t\sim \mathcal{N}(0, I)$, and $\sigma_t$ and $\alpha_t$ are scalar functions determined by the noise schedule that govern the rate of the backwards diffusion process. %

Despite the superior image quality afforded by DDPM, the sampling process sometimes involves $1\,000$ or more steps, which led to a variety of sampling schemes and distillation methods for speeding up inference~\cite{song2022denoising, kim2023consistency, song2023consistency, gu2023boot}.  
One of the most commonly used modifications to the sampling process is denoising diffusion implicit models (DDIM), %
which enables skipping steps in the backward process. 

Currently, the state-of-the-art for guided generation is achieved by models with classifier-free guidance~\cite{ho2022classifierfree}.  
However, since CPSample employs a classifier to prevent replication of its training data, it is more useful for us to review its predecessor, classifier-guided diffusion~\cite{score_based, dhariwal2021diffusion}. %
In classifier guided diffusion, a pretrained classifier $p_\phi(y\mid  x_t, t)$ assigns a probability to the event that $x_t = \sqrt{\alpha_t}x_0 + \sqrt{1-\alpha_t} \epsilon$ for some $x_0$ with label $y$. %
The sampling process for
classifier-guided DDIM is modified by \vspace{-1em}

\begin{align}
    \hat{\epsilon}_t = \epsilon_\theta(x_t) - \sqrt{1-\overline{\alpha}_t} \nabla_{x_t} \log p_\phi(y\mid x_t, t)\\
    x_{t-1} = \sqrt{\overline{\alpha}_{t-1}}\left(\frac{x_t - \sqrt{1-\overline{\alpha}_t} \hat{\epsilon}_t}{\sqrt{\overline{\alpha}_t}}\right) + \sqrt{1-\overline{\alpha}_t} \hat{\epsilon}_t.
\end{align}

Such a modification of the sampling procedure corresponds to drawing \( x_t \) from the joint distribution:
\begin{align} \label{joint_sampling}
    p_{\theta, \phi}(x_t, y \mid  x_{t+1}, t) = Z p_\theta(x_t \mid  x_{t+1}, t) p_\phi(y \mid  x_t, t)
\end{align}
where \( Z \) is a normalization constant. This formulation can be adapted for continuous-time models, but for discrete-time models, additional care must be taken to ensure accuracy (see Appendix \ref{classifier_guidance} for additional details).

\subsection{Privacy in Diffusion Models}

Differential privacy (DP) is generally considered to be the gold standard for protecting sensitive data.  
The formal definition of ($\epsilon$-$\delta$) differential privacy is as follows~\cite{dwork}:%

\begin{definition}[($\epsilon$-$\delta$)-Differential privacy]\label{diff_piv}
    Let $\mathcal{A}$ be a randomized algorithm that takes a dataset as input and has its output in $\mathcal{X}$.  If $D_1$ and $D_2$ are data sets with symmetric difference $1$, then $\mathcal{A}$ is $\epsilon$-$\delta$ differentially private if for all $S\subset \mathcal{X}$, 
    \begin{align}\label{diff_priv_equation}
        \mathbb{P}(\mathcal{A}(D_1)\in S) \leq \mathbb{P}(\mathcal{A}(D_2) \in S)e^{\epsilon} + \delta
    \end{align}
\end{definition}

DP ensures that the removal or addition of a single data point to the dataset does not significantly affect the outcome of the algorithm, thus protecting the identity of individuals within the dataset.
Existing DP diffusion models~\cite{dockhorn2023differentially, better_dpdm, latent_dpdm} achieve DP through DP stochastic gradient descent (DP-SGD)~\cite{dpsgd}, in which Gaussian noise is added to the gradients during training.

Though DP offers a formal guarantee that one's data is secure, imposing a DP constraint in practice severely compromises the quality of the synthetic images.  %
Therefore, researchers have largely resorted to demonstrating that models exhibit various relaxations of strict DP, such as Probably Approximately Correct DP (PAC-DP)~\cite{xiao2023pac} or other empirical metrics of privacy.
For instance, researchers measure the distance of generated images to their nearest neighbors in the training set and try to ensure that the number with similarity exceeding some threshold is small~\cite{daras2022soft, daras2023ambient}.
Usually, one computes similarity either via least squares or cosine similarity, given by \vspace{-1em}
\begin{align}\label{cosine_similarity}
    \frac{x^T \cdot \mathrm{C}(x)}{\lVert x\rVert \cdot \lVert \mathrm{C}(x)\rVert},
\end{align}
with $\mathrm{C}(x)$ denoting the nearest neighbor of $x$ among the training data.
Usually the cosine similarity is computed in a feature space rather than the raw pixel space~\cite{face_verification, XIA201539}.  %
For CIFAR-10, we observed that images with similarity scores above $0.97$ were nearly identical, whereas for CelebA, the threshold was approximately $0.95$.
For LSUN Church, images with similarity above $0.90$ were sometimes, though not always, nearly identical.  
While preventing exact replication of training data does not imply DP, a DP model will not exactly reveal members of its training data with high probability.
Therefore, preventing exact replication of training data recovers one of the desirable implications of DP. %
 
Ambient diffusion reduces similarity to the nearest neighbor by masking pixels during training and only scoring the model based on the visible pixels.
In this way, the model never has access to a full, uncorrupted image and is less likely to replicate full images, as it has not seen them \cite{daras2023ambient}.
The downside is that masking pixels, even at the relatively modest rate of $20\%$, leads to notable image quality degradation, measured via a notable increase in the FID score.
Moreover, ambient diffusion shifts the entire distribution of similarity scores towards lower similarity, whereas we should ideally prevent the generation of images with high similarity to the training data while leaving the rest of the distribution untouched. %
Tools have been developed to help compute nearest neighbors efficiently, since naïve pairwise comparisons are too computationally expensive.
In 2023, MetaAI developed the FAISS library for efficient similarity search using neural networks~\cite{douze2024faiss}, making this type of privacy metric possible to compute approximately in a reasonable amount of time.

Until recently, all attempts at enforcing privacy for diffusion models occurred during training.
In 2023, Xu~et~al.~\cite{xu2023pac} developed a method of classifier-guided sampling (PACPD) that has PAC privacy advantages over standard denoising.
For text-guided models, Somepalli~et~al.~\cite{somepalli2023understanding} developed a method of randomly changing influential input tokens to avoid exact memorization and Wen~et~al.~\cite{wen2024detecting} protected training data using a regularization technique on the classifier-free guidance network during training.  

\subsection{Membership Inference Attacks}

A third privacy measurement comes from membership inference attacks \cite{Dubinski_2024_WACV, pang2023blackbox, duan2023diffusion, wu2022membership}, whereby one tries to discern whether a given data point was a member of the training set for the model.  Robustness to membership inference attacks is implied by differential privacy.  Membership inference attacks against diffusion models usually hinge on observed differences in reconstruction loss or likelihood that come from overfitting.  In this paper, we will use a slight modification of the membership inference attack from \cite{matsumoto2023membership} given in Algorithm \ref{inference_attack} described in Appendix \ref{permutation_test}.

We repeat Algorithm \ref{inference_attack} for $x_1,...,x_m$ drawn first from the training data, and then from withheld test data.  If the resulting mean reconstruction error is significantly higher for test data than for training data, then we say that the diffusion model has failed the inference attack.

\section{Protecting Privacy During Sampling}

In this section, we address the problem of training data replication in diffusion models, which poses significant privacy risks. One common solution to this problem is rejection sampling, where samples that closely resemble training data are discarded. However, rejection sampling has several shortcomings: it is computationally expensive, inefficient, and does not provide protection during the sampling process itself, but only in the final output.  Moreover, in extreme cases of mode collapse, one must generate dozens of images before generating original content when rejection sampling.  %

To overcome these limitations, we introduce CPSample, a method that reproduces some of the benefits of rejection sampling without the need for resampling. CPSample integrates classifier guidance into the sampling process to steer the generation away from the training data. By overfitting a classifier on random binary labels assigned to the training data, we use this classifier during sampling to adjust the generated images, thus reducing the likelihood of replicating training data while maintaining image quality.

Let $B_\delta(x)$ denote the ball of radius $\delta$ around $x$ in a given metric space.  Explicitly, for a training dataset $D$, $\delta>0$ that is not too large, and $\epsilon>0$, our goal is to produce a diffusion model that generates data $\Tilde{x}$ such that $\mathbb{P}\left(\Tilde{x} \in  \bigcup_{x\in D} B_\delta(x)\right) < \epsilon$ while compromising image quality as little as possible in the process.

\subsection{Sampling Method} \label{sampling_method}
The first step in CPSample is to train a network that can provide information about how likely a sample $x_t$ is to turn into a member of the training data at the end of the denoising process.  For this task, we use a classifier trained to memorize random binary labels assigned to the training data, which is possible for a sufficiently over-parameterized network relative to the number of points being memorized \cite{random_labels_orig}.  %
During the denoising process, whenever the classifier predicts a label $y \in \{0,1\}$ for $x_t$ with probability greater than $1-\alpha$, we perturb $x_{t-1}$ towards the opposite label using classifier guidance.  %
Without loss of generality, if the classifier predicts the label $1$ with high probability, we employ classifier guidance to adjust the sampling process. Specifically, we aim to sample from a modified distribution that reduces the likelihood of the generated sample being close to the training data. We employ classifier guidance to perturb the denoising process to sample from the conditional distribution $p_{\theta, \phi}(x_{t-1}\mid x_t, t, y=0)$. This approach ensures that the modified sampling process leads to a lower density near points in the training data that $x_t$ might have otherwise converged to.

    To state our procedure more precisely, let $\epsilon_\theta(\cdot, \cdot)$ be the denoiser. Note that the classifier is trained once on the training data and not during each sample generation. The sampling process is then modified in the following steps:    %
\begin{enumerate}
    \item Randomly assign Bernoulli$(0.5)$ labels to each member of the training data, and let $B\in \{0,1\}^n$ index these random labels. Train a classifier $p_\phi(y\mid  x_t, t)$ to predict these labels. Here, $x_t$ is generated by corrupting the training data $x_0$ with noise: $x_t = \sqrt{\alpha_t}x_0 + \sqrt{1-\alpha_t}\epsilon$ for $\epsilon \sim \mathcal{N}(0, I)$ and $t \in \{0, ..., T\}$.
    \item Set a tolerance threshold $0<\alpha<0.5$ and a scale parameter $s$.  Let $p_\phi(y\mid x_t, t)$ be the probability assigned to the label $y$ by the classifier $p_\phi(y\mid  x_t,t)$.  Sample $x_T\sim \mathcal{N}(0, I)$.  For $t\in \{T,....,1\}$, if $p_\phi(y=0\mid x_t,t) < \alpha$, replace $\epsilon_\theta(x_t, t)$ with $$\hat{\epsilon}_{\theta, \phi}(x_t,t) = \epsilon_\theta(x_t,t) - s\sqrt{1-\overline{\alpha}_t}\cdot \nabla \log ( p_\phi(y=0\mid x_t, t)).$$  If $p_\phi(y=1\mid x_t,t) < \alpha$, replace $\epsilon_\theta$ with $$\hat{\epsilon}_{\theta, \phi}(x_t,t) = \epsilon_\theta(x_t,t) - s\sqrt{1-\overline{\alpha}_t}\cdot \nabla \log (p_\phi(y=1\mid x_t,t)).$$
    Otherwise, we leave the sampling process unchanged.
\end{enumerate}

Though the choice of random labels for the classifier initially may seem counter-intuitive, it has several advantages over other approaches. If we used labels corresponding to real attributes of the data, the classifier would influence the content of the generated images in ways that could compromise their authenticity and diversity. This is because the guidance would push the generated images towards or away from specific attributes, altering the intended distribution. The perturbation applied by the gradient of the log probability in CPSample moves the generated images away from regions where they can be easily classified as similar to the training data. This method is more effective than adding random noise, which would require a significant amount of noise to achieve the same effect, thus degrading image quality.

Moreover, without a classifier, detecting when the denoising process is likely to recreate a training data point would be challenging. The classifier provides a mechanism to identify and steer image generation away from such points. If we trained the classifier to distinguish between train and test data and guided the diffusion process towards the test data, we might inadvertently replicate or reveal aspects of the test data.

Unlike past training-based methods of privacy protection, once we have trained the classifier, we can adjust the level of protection by tuning the hyperparameters $s$ and $\alpha$ without necessitating retraining of the classifier or denoiser.  In ambient diffusion and DPDM, if one realizes after training that one needs a higher or lower level of privacy, one must retrain the entire diffusion model from scratch.

\subsection{Theory}\label{theory}

In this section, we show that CPSample functions similarly to rejection sampling when preventing exact replication of the training images.  We work under the following assumptions:

\begin{assump} \label{lipschitz}Suppose that the classifier $p_\phi(y\mid x, t)$ has Lipschitz constant $L$ in the argument $x$ with respect to a metric $d(\cdot, \cdot):\chi\times \chi \rightarrow \mathbb{R}_{\geq 0}$, where $\chi$ denotes the image space. \end{assump}
\begin{assump}\label{accuracy} Let $y_i$ be the random label assigned to $x_i \in D$, where $D$ is the training data.  
Suppose that the classifier $p_\phi$ is trained sufficiently well such that for $\gamma < 1$, \begin{equation} \sum_{x_i \in D} \mathbb{P}\left(p_\phi(y_i\mid  \sqrt{\alpha_t}x_i + \sqrt{1-\alpha_t}\epsilon, t) \notin (1-\kappa, 1] \right) < \gamma \end{equation} for all $x_i\in D$, $t\in [0, T]$ and $\epsilon \sim \mathcal{N}(0, I)$.  \end{assump}

\begin{assump}\label{gen}
    Suppose that CPSample generates data $\Tilde{x}$ such that $\lambda < p_\phi(y\mid \Tilde{x}, 0) < 1-\lambda$ with probability greater than $1-\nu$, where we are able to govern $\nu$ and $\lambda$ by adjusting $s$ and $\alpha$ in Section~\ref{sampling_method}.
\end{assump}

In Assumption~\ref{lipschitz} the constant $L$ can be difficult to evaluate, but the assumption holds for neural network classifiers.  Methods exist that can bound the local Lipschitz constant around the training data \cite{local_lipschitz}, which one can use to strengthen the guarantees of Lemma \ref{rejection_sampling}.
Assumptions \ref{accuracy} and \ref{gen} hold well empirically.  
We were able to train our classifier to have a cross-entropy loss below $0.05$ in the experiments from Sections~\ref{similarity_reduction_subset} and \ref{stable_diffusion}.  Moreover, during sampling, we observed that CPSample had control over the quantity $p_\phi(y\mid x_t, t)$.  An example is given in Figure~\ref{probability_control}.  

Given these assumptions, we can demonstrate the following simple lemma, which links the behavior of CPSample to that of a rejection sampler without requiring expensive comparisons to the training data set.  A proof can be found in Appendix~\ref{classifier_guidance}.

\begin{lem}\label{rejection_sampling}
    Under the above assumptions, choose $\epsilon>0$ and $0< \delta < \frac{\frac{1}{2}- \kappa}{L}$.  %
    Setting $\nu = \epsilon$ and $\lambda = \kappa + L\delta$, when drawing a single sample, with probability greater than $(1-\epsilon)(1-\gamma)$, CPSample generates an image that lies outside of $S= \bigcup_{x\in D} B_\delta(x)$ in the metric space defined by $d$.  

\end{lem}

Note that the ability to control $\mathbb{P}\left(\Tilde{x} \in \bigcup_{x\in D} B_\delta(x)\right)$ gives the same guarantee offered by rejection sampling.  However, in extreme instances of mode collapse such as those exhibited by Stable Diffusion in Section~\ref{stable_diffusion}, one might have to resample hundreds of times to generate original images, making standard rejection sampling highly inefficient.  CPSample is able to produce original images without this high level of inefficiency.

\section{Empirical Results}\label{empirical}

\begin{figure}
    \centering
    \includegraphics[width=0.99\linewidth]{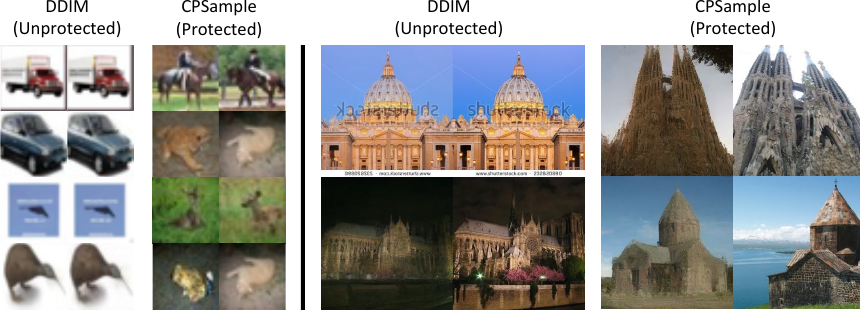}
    \caption{Generated image and most similar training image pairs for DDIM sampling and CPSample with $\alpha = 0.001, s=1$ on CIFAR-10 (left) and $\alpha = 0.1, s = 10$ on LSUN Church (right).  For each pair, the image on the left is the generated sample and the one on the right is its nearest neighbor in the training set. These are the four examples out of $21\,000$ images on CIFAR-10 and two out of $1\,700$ images on LSUN Church that have the highest similarity scores with their nearest neighbor.}
    \label{fig:sim_red_cifar_church}
\end{figure}
We run three distinct sets of experiments to demonstrate the ways in which CPSample protects the privacy of the training data.  First, we statistically test the ability of CPSample to reduce similarity between generated data and the training set for unguided diffusion.  Then, we demonstrate that CPSample can prevent Stable Diffusion from generating memorized images.  Finally, we measure robustness against membership inference attacks.  Hyperparameters in all empirical tests were chosen to maximize image quality while eliminating exact matches.  %
 
\begin{figure}[t!]
    \centering
    \includegraphics[width=0.95\linewidth]{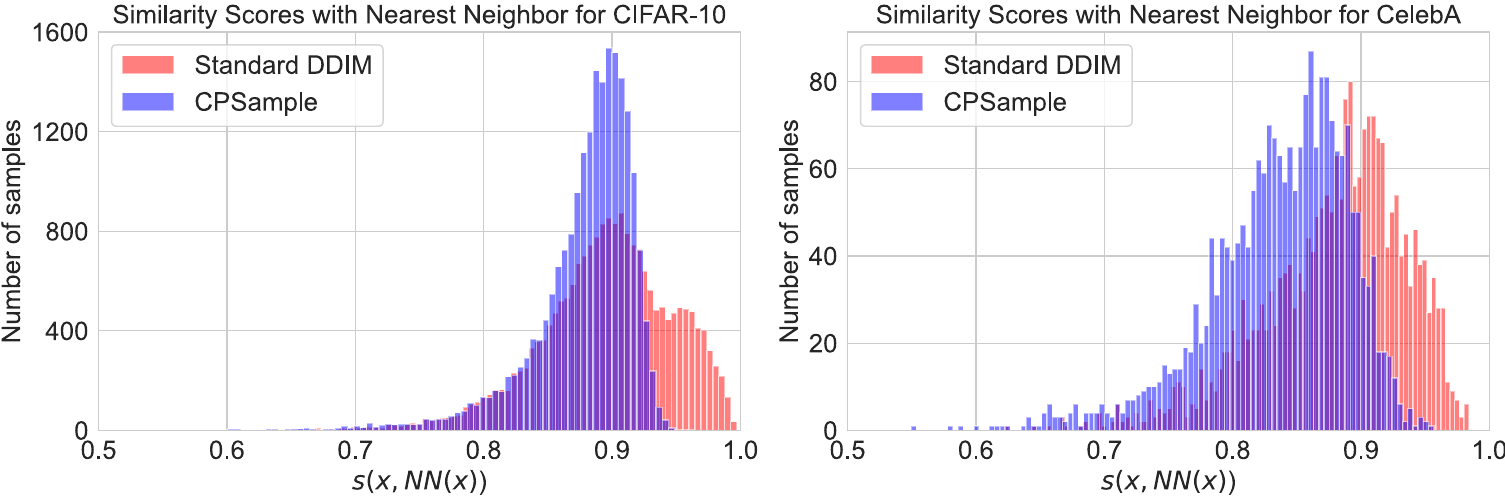}
    \caption{\label{CIFAR-similarity} Cosine similarity in feature space between generated images and their nearest neighbor in the fine-tuning data set for standard DDIM sampling (red) and CPSample (blue) on CIFAR-10 ($\alpha = 0.001, s=1$) and CelebA-64 $(\alpha =0.001, s = 1000)$.  Similarity scores were computed for $21\,000$ generated samples for CIFAR-10 and $8\,000$ images for CelebA.  Note that standard DDIM exhibits many more samples with similarity scores exceeding the thresholds from Table \ref{similarity_numeric}.  } 
\end{figure}

\begin{table}[b]
  \centering
  \caption{\label{similarity_numeric}Reduction in cosine similarity between generated images and nearest neighbor in fine-tuning data.}\vspace{.5em}
  \setlength{\tabcolsep}{4pt}
  \begin{threeparttable}
       \begin{tabular}{cccccccc}
    \toprule
         Dataset & FT Steps & $\alpha$ & Scale & Threshold & DDIM& CPSample & p-value\tnote{1}\\\midrule CIFAR-10 & 150k& 0.001 & 1 & 0.97 & 6.25$\%$ & 0.00 $\%$ & <0.0001 \\ 
         CelebA & 650k & 0.001 & 1000 & 0.95 & 12.5$\%$ & 0.10$\%$& <0.0001 \\ LSUN Church & 455k & 0.1 & 10 & 0.90 & 0.73$\%$& 0.04$\%$ & 0.013 \\\bottomrule
    \end{tabular}%
    \begin{tablenotes}
        \item[1] $p$-values were computed using a log rank test for $H_0$: CPSample did not reduce the fraction of images with similarity score exceeding the threshold.
    \end{tablenotes}
  \end{threeparttable}
\end{table}%
\subsection{Similarity Reduction}\label{similarity_reduction_subset}

We generate images using DDIM with CPSample and $1000$ denoising steps.  The nearest neighbor to each generated image was found using Meta's FAISS model \cite{douze2024faiss}.  Similarity between two images is measured by cosine similarity in a feature space defined by FAISS.  A similarity score exceeding 0.97 often indicates nearly-identical images for CIFAR-10.  For CelebA and LSUN Church, the thresholds lie around 0.95 \cite{daras2023ambient} and $0.90$ respectively.  Note that a cosine similarity score above the thresholds given is a necessary but not sufficient condition for images to look very alike.  %
  To ensure that we could observe a larger number of images with similarities exceeding our thresholds, we fine-tuned the models using DDIM \cite{song2022denoising}
  on a subset of the data that consisted of 1000 images, as was done in \cite{daras2023ambient}.  This modification allows us to statistically test the efficacy of CPSample without the large number of samples required to do hypothesis testing on rare exact replication events. 
 After fine-tuning, up to $12.5\%$ of the images produced by unprotected DDIM were nearly exact replicates of the fine-tuning data.  One can see from Table \ref{similarity_numeric} that CPSample significantly reduces the fraction of generated images that have high cosine similarity to members of the fine-tuning set.  One can see histograms of the similarity score distribution with and without CPSample in Figures \ref{CIFAR-similarity} and \ref{fig:lsun_church_reduction}. Figures \ref{fig:sim_red_celeba} and \ref{fig:sim_red_cifar_church} show the most similar pairs of samples and fine-tuning data points.  %
 Uncurated images generated from CPSample can be found in Appendix \ref{empirical_dump}.
While CPSample effectively reduces the similarity between generated images and the training data, our experiments indicate that CPSample achieves minimal degradation in quality compared to previous methods \ref{fid_tab}.

\begin{figure}[h!]
    \centering
    \begin{minipage}{0.38\textwidth}
        \centering
        \includegraphics[width=\linewidth]{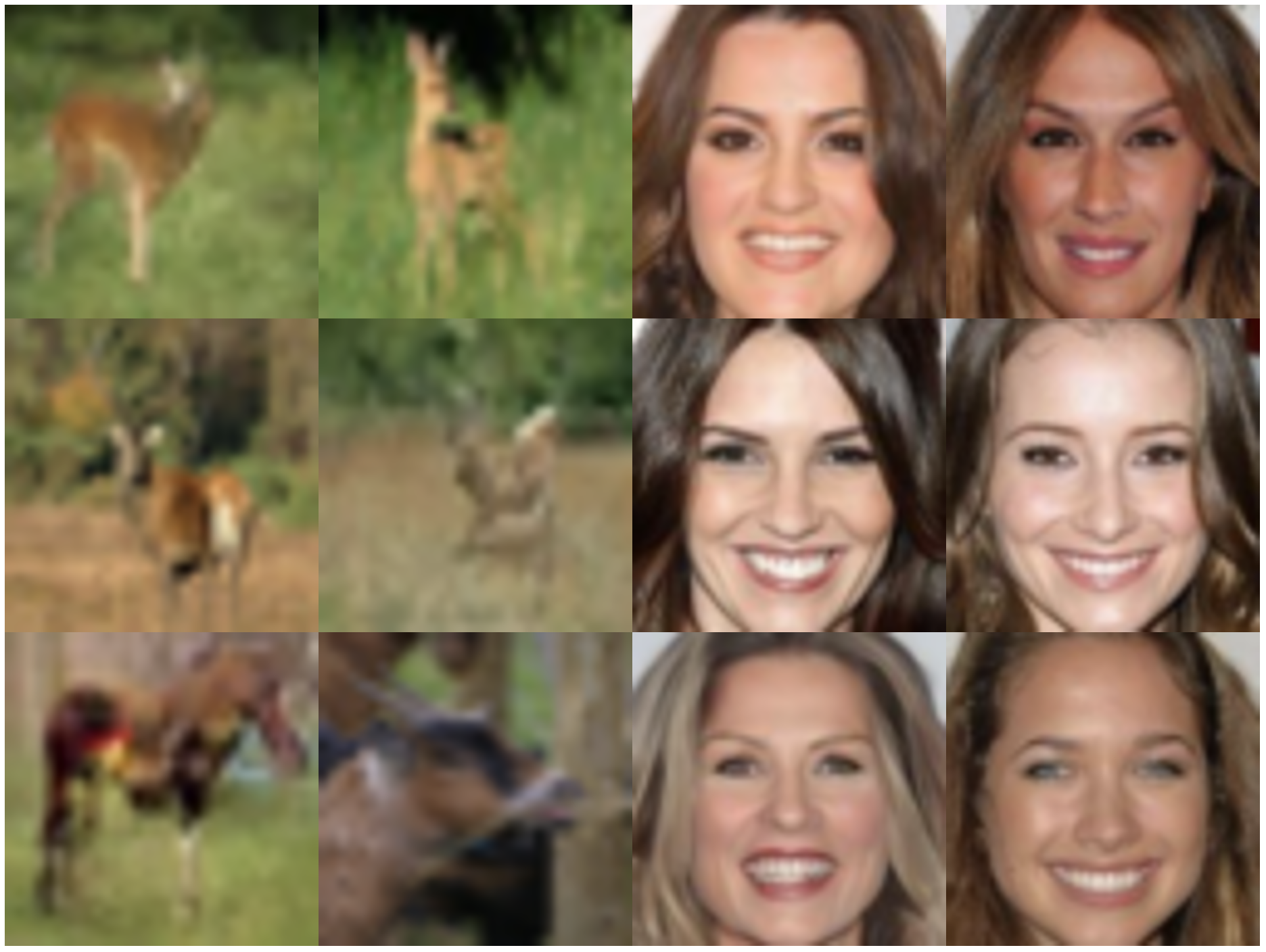} 
        \caption{The generated and real images with the highest similarity for CIFAR-10 (left) and CelebA (right) out of $50\,000$ samples used to compute FID score.}
     \label{most_similar_full}
    \end{minipage}\kern3em
    \begin{minipage}{.50\textwidth}
        \centering
        \includegraphics[width=\linewidth]{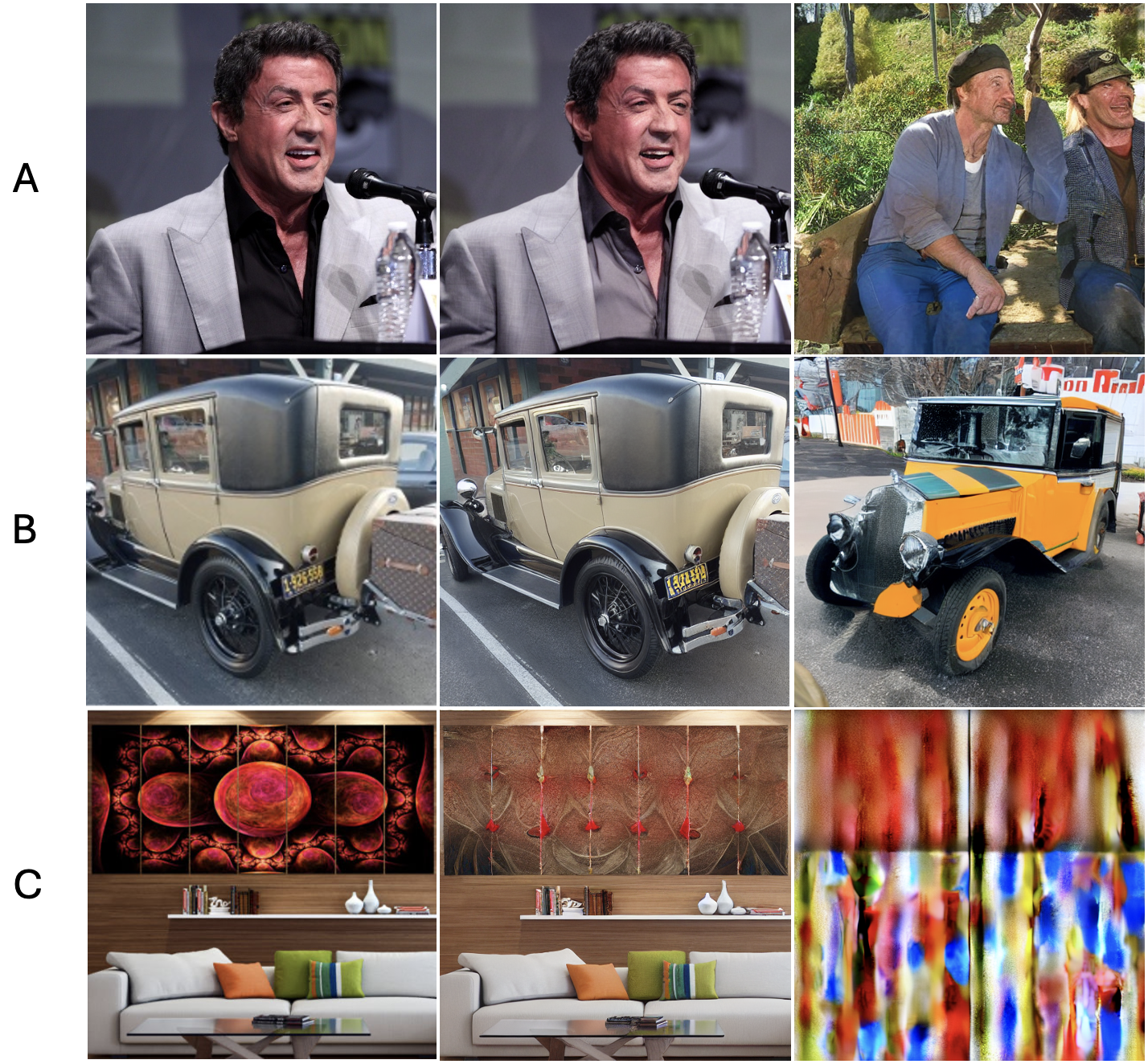}
        \caption{Selected examples for Stable Diffusion:
        original image (left), image generated from a similar caption by Stable Diffusion v1.4 (center), image generated with CPSample (right).
        }
        \label{fig:sd}
    \end{minipage}
\end{figure}

\subsection{Stable Diffusion}\label{stable_diffusion}

As a second demonstration of CPSample, we present evidence that CPSample can prevent well-known examples of mode collapse in near-verbatim attacks against Stable Diffusion \cite{webster2023reproducible, wen2024detecting}.  We create a small dataset of commonly reproduced images \cite{somepalli2023understanding} and include other images from the LAION dataset depicting the same subjects.  We ensure that this dataset contains no duplicates.  %
In this more targeted application, CPSample can prevent exact replication when used with the right hyperparameters. See Figure~\ref{fig:sd} and Table \ref{sd_details} for more details. Though CPSample does not provide as robust protection in this setting as in~\cite{somepalli2023understanding, wen2024detecting}, these results still show the potential of CPSample for data protection in text-guided diffusion models.  The methods developed in~\cite{somepalli2023understanding, wen2024detecting} do not apply for unguided diffusion models.

  \begin{table}[h!]
  \centering\vspace{1em}
   \caption{\label{sd_details} Details of generation on Stable Diffusion.}%
       \begin{tabular}{c>{\raggedright}p{4cm} >{\raggedright}p{4cm} ccc}
    \toprule
        Image & Original caption & Modified caption  & $\alpha$ & scale & guidance \\
    \midrule
      A &   ``Rambo 5 and Rocky Spin-Off - Sylvester Stallone gibt Updates"
  &  ``Rocky and Rambo Spin-Off - Sylvester Stallone gibt Updates"    & 0.5 & 2000 & 1.5  \\
    B& ``Classic cars for sale"
  &  ``Classic car for sale"    & 0.3 & 100 & 1.5  \\    C& ``Red Exotic Fractal Pattern Abstract Art On Canvas-7 Panels"
  &  ``Red Exotic Fractal Pattern Abstract Art On Canvas-7 Panels"    & 0.5 & 2000 & 1.5  \\
    \bottomrule
    \end{tabular}%
  \end{table}

\subsection{Membership Inference Attacks}\label{white_box_inference_attacks}

We also assess CPSample ability to protect against membership inference attacks.  Following Algorithm \ref{inference_attack}, we compute the mean reconstruction error for the train and test datasets and determine whether there is a statistically significant difference.  
To evaluate resistance to inference attacks, we use a model trained on all $50\,000$ CIFAR-10 training images.  We compare reconstruction loss on these $50\,000$ training images to reconstruction loss on the $10\,000$ withheld test samples included in the CIFAR-10 data set. We juxtapose the difference in reconstruction loss between these two datasets for both CPSample, with a classifier $c_\phi$ trained on all of the CIFAR-10 training data with random labels, and standard DDIM sampling. We demonstrate resistance to inference attacks for $\alpha\in \{0.5, 0.25, 0.001\}$ over about $8000$ images from each of the train and test sets.  The $p$-values in this experiment are based on a two-sample, single-tailed $Z$-score that tests the null hypothesis ``the average training reconstruction loss is less than or equal to the average test reconstruction loss."  Let $n$ be the number of training data points and $m$ be the number of test data points sampled.  The test statistic is \[\frac{\mu_\textrm{test}-\mu_{\textrm{train}}}{\sqrt{V_{\textrm{test}}/m+V_{\textrm{train}}/n}}.\] The variable $V$ indicates the variance and $\mu$ indicates the mean. 
 In our case, failure to reject the null indicates success for CPSample.  %

We observe that in our experiments, a very low value of $\alpha$ leads to a higher $p$-value, which is counter intuitive.  We suspect that this occurs because when $\alpha$ is small, it results in a more targeted application of CPSample, driving the loss up exclusively around the training data points.  However, for all values of $\alpha$ between $0$ and $0.5$, one is not able to conduct a conclusive membership inference attack against CPSample. We provide a second black-box membership inference attack based on permutation testing in Appendix \ref{permutation_test}.

\subsection{Quality Comparison} \label{fid}
As mentioned, other methods of privacy protection suffer from severe degradation of quality, as measured by FID score.  Here, we provide an FID score comparison between CPSample model finetuned on subsets of CIFAR-10 and CelebA, and existing methods of privacy protection.  We exhibit FID scores for unconditional generation of CIFAR-10 and CelebA in Table \ref{fid_tab}.  The images with the highest similarity to the training set, as determined by FAISS, can be found in Figure~\ref{most_similar_full}. 
 We chose $\alpha$ and $s$ by finding the least aggressive settings that completely prevent exact replication of the training data. For more extensive results on how $\alpha$ and $s$ affect FID score, see Table \ref{fid_effects}. %

\begin{table}
\begin{minipage}[t]{0.47\textwidth}
\centering
  \caption{Difference in mean reconstruction error between train and test data for CIFAR-10.  
  }
  \setlength{\tabcolsep}{1.5pt}
    \scalebox{.95}{
  \begin{threeparttable}
       \begin{tabular}{cccc}
    \toprule
         Method & Test statistic  & p-value \\
    \midrule
    DDIM  &   138   &   $\approx$ 0      \\
        Ambient (Corruption $0.2$) & 0.141 & 0.44 \\
    Ambient~(Corruption 0.8) & -0.024 & 0.51 \\
    \midrule
    CPSample ($\alpha =0.5$)   &  0.59     & 0.28  \\
    CPSample ($\alpha=0.25$)  &   0.23 & 0.41 \\
    CPSample ($\alpha=0.001$) &    -0.86 & 0.81  \\
    \bottomrule
    \end{tabular}%
  \end{threeparttable}}
\end{minipage}
\hfill
\begin{minipage}[t]{0.48\textwidth}
\centering
  \caption{\label{fid_tab} FID Score comparison on the CIFAR-10 and CelebA datasets.}\vspace{-.5em}
    \setlength{\tabcolsep}{1pt}
    \scalebox{.95}{
    \begin{tabular}{lcc}
    \toprule
          & \multicolumn{2}{c}{FID} \\
              \cmidrule{2-3}
          & CIFAR-10 & CelebA \\
    \midrule
        DDIM & 3.17 & 1.27\\
    \midrule
    Ambient (Corruption 0.2) & 11.70 & 25.95\\
    DPDM ($\epsilon=10$)  & 97.7 & 78.3\\
    DP-Diffusion ($\epsilon =10$)& 9.8 & -\\
    DP-LDM ($\epsilon =10$) & 8.4 & 16.2 \\ 
    \midrule
    CPSample $(\alpha = 0.001, 0.05)$ &  \textbf{4.97} & \textbf{2.97}\\
    \bottomrule
    \end{tabular}}%
  \label{tab:fid-cifar}%
\end{minipage}
\end{table}

\section{Limitations}\label{limitations}
\vspace{-.25em}

While CPSample prevents training data replication well, it comes with the drawback that it can be difficult to train a classifier on binary random labels for large data sets, making CPSample better suited to protecting small to moderately sized datasets. Nevertheless, we were able to successfully protect datasets of up to $180\,000$ images.  With larger datasets, it could become difficult to obtain a classifier that has high accuracy while maintaining a reasonably low Lipschitz constant without introducing Lipschitzness regularizations. When training only on a few images, we noticed instability in the classifier for Stable Diffusion that made it difficult to guarantee complete protection across seeds and at higher levels of classifier-free guidance, which could be attributed to non-pretrained classifier models. Additional research on regularization could help to alleviate this limitation.  %

If there are multiple exact repeats in the training set, CPSample could struggles to protect the training data if opposite labels were assigned to the repetitions, though it still can perform well when there are multiple near-repeats. Research surrounding removing repeats or randomly transforming the data to prevent exact repeats would likely improve CPSample's performance.  Repeats could also be strategically grouped to have the same label.

Finally, given the difficulty of measuring the Lipschitz constant of a neural network, it may be difficult to give practical bounds on the protection rate of CPSample.  
In practice, we observe stronger protections than the formal guarantees provide.

\vspace{-.5em}
\section{Conclusion}
\vspace{-.25em}
We have presented a new approach to prevent memorized images from appearing during inference time.  Our method is applicable to both guided and unguided diffusion models.  Unlike previous methods intended to protect privacy of unguided diffusion models, CPSample does not necessitate retraining the denoiser.  Moreover, after training the classifier, one can adjust the level of protection enforced by CPSample without further training. We have shown theoretically that our method behaves similarly to rejection sampling without necessitating resampling.  Finally, we have provided empirical evidence with rigorous statistical testing that our method is effective in unguided settings.  We have also given examples in which CPSample was able to prevent extreme instances of mode collapse in Stable Diffusion.  Despite its efficacy at preventing replication of training images, CPSample has little negative impact on image quality.

\vspace{-.25em}
\subsection*{Broader Impact}\label{broader_impacts}
\vspace{-.25em}
Both guided and unguided diffusion models are prone to reproducing their training data and consequently violating copyright or privacy. 
 As diffusion models become more widespread in entertainment and commercial settings, these issues have the potential to cause real harm.  We explore a new avenue to prevent replication and promote robustness to membership inference attacks through early detection and perturbation.  When deployed responsibly, CPSample has the potential to prevent the majority of instances in which exact replication occurs.

\printbibliography

\clearpage
\appendix

\section{Proofs}\label{classifier_guidance}

\paragraph{Details of classifier guidance} For completeness, we include a derivation of the classifier-guidance introduced in \cite{dhariwal2021diffusion}.  

During the conditional denoising process, one should sample $x_{t-1}$ from the conditional distribution \begin{equation} \mathbb{P}(x_{t-1}\mid x_t, y) = \frac{\mathbb{P}(x_{t-1}, x_t, y)}{\mathbb{P}(x_t, y)} = \frac{\mathbb{P}(x_{t-1}\mid x_t) \mathbb{P}(y\mid x_{t}, x_{t-1})}{\mathbb{P}(x_t, y)}. \end{equation} 
One can show that $\mathbb{P}(y\mid x_t, x_{t-1}) = \mathbb{P}(y\mid x_{t-1})$ (see \cite{dhariwal2021diffusion} for details).  The denominator $\mathbb{P}(x_t, y)$ is intractable and does not depend on $x_{t-1}$.  Therefore, we write this term as $Z$.  To get an estimate of the probability $\mathbb{P}(y\mid  x_{t-1})$, we train a classifier of the form $p_\phi(y\mid x_{t-1})$.  Thus, we should estimate the conditional probability $\mathbb{P}(x_{t-1}\mid x_t, y)$ via \begin{equation} p_{\theta, \phi}(x_{t-1}, x_{t}, y) = Zp_\theta(x_{t-1}\mid x_{t})p_\phi(y\mid x_{t-1}).\end{equation}

In continuous time, we can write $p(x_t, y) = p(x_t)p(y\mid x_t)$, and the score function is:

\begin{align} \label{joint_score}
    \nabla_{x_t} \log(p_\theta(x_t)p_\phi(y\mid x_t)) &= \nabla_{x_t} \log p_\theta(x_t) + \nabla_{x_t} \log p_\phi(y\mid x_t) .\end{align}

The network $\epsilon_\theta(x_t,t)$ predicts the noise added to a sample, which can be used to derive the score function \[\nabla_{x_t} \log p_\theta(x_t, t) = -\frac{1}{\sqrt{1-\overline{\alpha}_t}} \epsilon_\theta(x_t,t).\]  Substituting this into \eqref{joint_score}, we get 
\begin{equation} -\frac{1}{\sqrt{1-\overline{\alpha}_t}} \epsilon_\theta(x_t) + \nabla_{x_t} \log p_\phi(y\mid x_t).
\end{equation}

This leads to a new prediction for \[\hat{\epsilon}_\theta(x_t) = \epsilon_\theta(x_t) -\sqrt{1-\overline{\alpha}_t}\nabla_{x_t} \log p_\phi(y\mid x_t).\]  The conditional sampling then follows in the same manner as standard DDIM with $\epsilon_\theta$ replaced by $\hat{\epsilon}_\theta$.

\paragraph{Proof of Lemma \ref{rejection_sampling}}
\begin{proof}
    Let $x' \in B_\delta(x_0)$, where $x_0 \in D$ is assigned the random label $y$. %
    By the definition of Lipschitz, we have that \[|p_\phi(y\mid x_0, t) - p_\phi(y\mid x', t) | < Ld(x_0, x').\]  Since $p_\phi(y\mid x_0, 0) > 1-\kappa$, it follows that \begin{align*}
        p_\phi(y\mid x', 0) &= p_\phi(y\mid x_0, 0) - p_\phi(y\mid x_0, 0) + p_\phi(y\mid x', 0) \\ 
        &= p_\phi(y\mid x_0, 0) - (p_\phi(y\mid x_0, 0) - p_\phi(y\mid x', 0)) \\ 
        &\geq p_\phi(y\mid x_0, 0) - |p_\phi(y\mid x_0, 0) - p_\phi(y\mid x', 0)|
        \\ & \geq p_\phi(y\mid x_0, 0) - Ld(x_0, x')
        \\ & \geq p_\phi(y\mid x_0, 0) - L \delta \\ &\geq 1-\kappa -  L\delta \\ &= 1-\lambda.
    \end{align*}

By assumption, CPSample generates samples $\Tilde{x}$ with $\lambda < p_\phi(y\mid \Tilde{x}) < 1-\lambda$ with probability at least $1-\nu$.  Because all points $x'\in S$ have $p_\phi(y\mid x', 0) \in [0, \lambda] \bigcup [1-\lambda, 1]$ with probability at least $1-\gamma$, CPSample must generate its samples $\Tilde{x}\in \chi \setminus S$ with probability at least $(1-\epsilon)(1-\gamma)$.
\end{proof}

\begin{figure}[H]
    \centering
    \includegraphics[scale= 0.5]{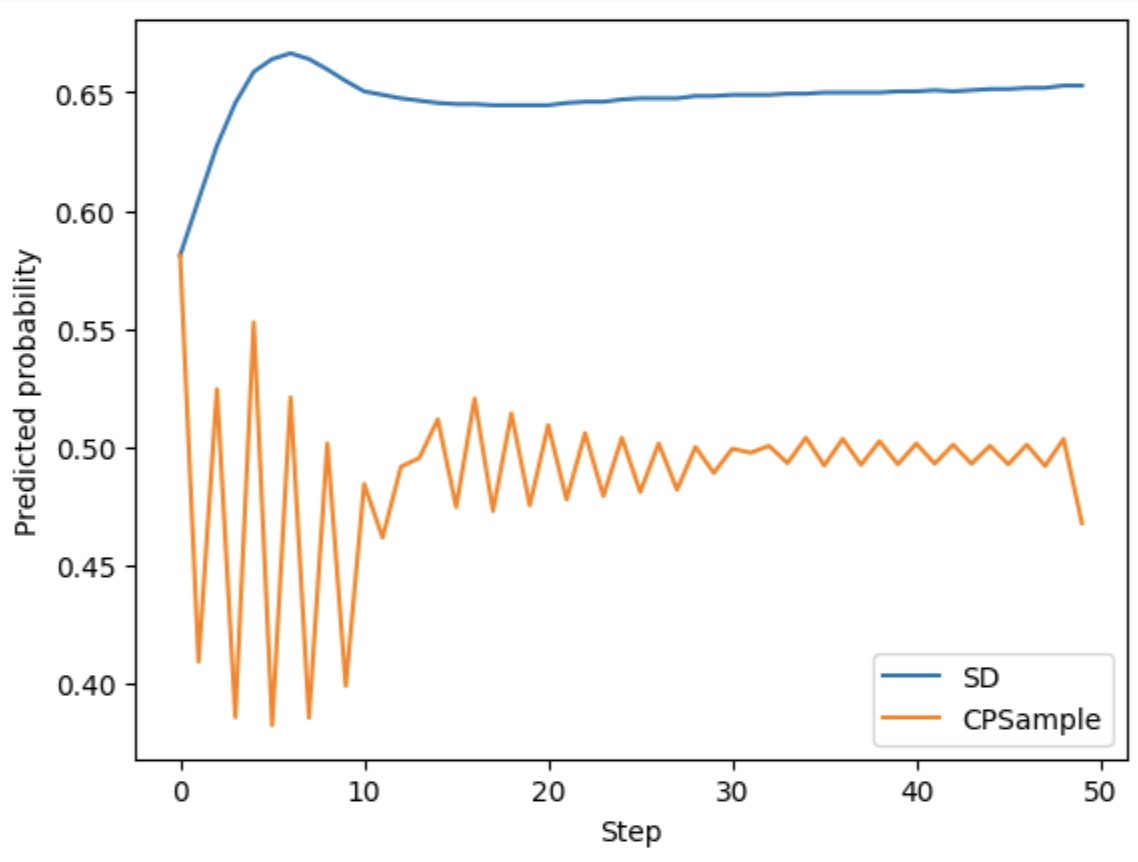}
    \caption{CPSample is able to generate images with $p_\phi(y\mid \Tilde{x}, 0) \in (\lambda, 1-\lambda)$.  This example shows the probability $p_\phi(y=1\mid x_t,t)$ during the generation process with Stable Diffusion guided by the caption ``Rambo 5 and Rocky Spin-Off - Sylvester Stallone gibt Updates."  Note that a higher step indicates a later point in the denoising process.  In this example, Stable diffusion exactly replicated the memorized image of Stallone, whereas CPSample $(\alpha = 0.5, s = 2000)$ produced an original image.}
    \label{probability_control}
\end{figure}

\section{Class Guided Diffusion} \label{guided}

As a final experiment, we implement CPSample alongside classifier-free guidance for CIFAR-10 to ensure that CPSample does not cause frequent out-of-category samples.  The models used for guided diffusion were smaller, so the image quality is naturally lower.

\begin{figure}[H]
    \centering
    \includegraphics[scale= 0.7]{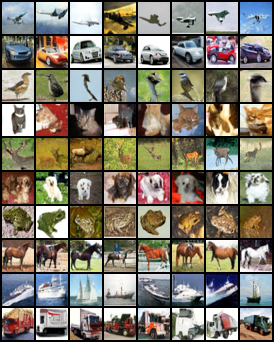}
    \caption{Uncurated samples using classifier-free guidance on CIFAR-10.  The image in the position second row, third column from the top left is a near-exact replica of a member of the training data.  }
\end{figure}

\begin{figure}[H]
    \centering
    \includegraphics[scale= 0.7]{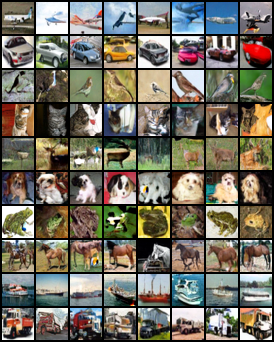}
    \caption{Uncurated samples using CPSample $(\epsilon = 0.1, s = 10)$ along with classifier-free guidance on CIFAR-10. Note that although CPSample slightly reduces image quality, it does not cause out-of-category samples.}
\end{figure}

\section{Training details}\label{training_details}

\paragraph{Training classifiers.} For training the classifier, we randomly selected subsets of 1000 images each from the CIFAR-10, CelebA, and LSUN Church datasets, on which we trained the classifier from scratch. The architecture of our classifier is a modified version of the U-Net model. We retained key components of the U-Net~\cite{ronneberger2015unet} model structure, including the timestep embedding, multiple convolutional layers for downsampling, and middle blocks. The output from the middle blocks underwent processing through Group Normalization, SiLU~\cite{elfwing2017sigmoidweighted} activation layers, and pooling layers before being fed into a single convolutional layer, yielding the classifier's output. Parameters for layers identical to the standard U-Net were consistent with those used to pretrain the DDIM model on these datasets. Additionally, akin to the pretraining of DDIM, we incorporated Exponential Moving Average during training to stabilize the training process. The training of each classifier model was conducted using 4 NVIDIA A4000 GPUs with 16GB of memory.  For subsets of $1000$ images, the classifier took only hours to train.  For larger datasets consisting of $60,0000-160,000$ data points, the classifier took up to $1$ week to train.  By comparison, retraining a diffusion model to be differentially private or using the method presented in \cite{daras2023ambient} can take weeks or months depending on the data set.  

\paragraph{Fine-tuning pretrained denoiser model on subsets.} For fine-tuning the pretrained denoiser model on subsets, we commenced with the 500,000-step pretrained checkpoints available for the denoiser DDIM model. Fine-tuning was performed on subsets of 1000 images each from the CIFAR-10, CelebA, and LSUN Church datasets until the model began generating data highly resembling the respective subsets. The number of training steps varied across different models, and specific details regarding the fine-tuning process can be found in Table~\ref{tab:training}. Throughout the fine-tuning process, hyperparameters remained consistent with those used during the pretraining phase. We employed 2 NVIDIA A5000 GPUs with 24GB of memory for fine-tuning each model on the subsets.

\begin{table}[h]
\small
\centering
\caption{Training Parameters \& Steps}\setlength{\tabcolsep}{2pt}
\begin{tabular}{lcccccc}
\toprule
 & \textbf{Batch Size} & \textbf{LR} & \textbf{Optimizer} & \textbf{EMA Rate} & \textbf{Classifier Steps} & \textbf{Fine-tune Steps}  \\
\midrule
\textbf{CIFAR-10} & 256 & 2e-4 & Adam & 0.9999 & 560,000 & 110,000 \\
\textbf{CelebA} & 128 & 2e-4 & Adam & 0.9999 & 610,000 & 150,000 \\
\textbf{LSUN Church} & 8 & 2e-5 & Adam & 0.999 & 1250,000 & 880,000 \\
\bottomrule
\end{tabular}
\label{tab:training}
\end{table}

\section{Evaluation Details}\label{eval_details}

\paragraph{Numerical stability} For the purposes of numerical stability, we slightly modified the sampling process described in Section \ref{sampling_method}.  We noticed in earlier iterations of our method that very small numbers of images were becoming discolored or black because in float16, the classifier was predicting probabilities of 0.0000 or 1.0000 for the random label 1, causing the logarithm to blow up.  To fix this in practice, we do the following. Sample $x_T\sim \mathcal{N}(0, I)$.  For $t\in \{T,....,1\}$, if $p_\phi(y=0|x_t,t) < \alpha$, replace $\epsilon_\theta(x_t, t)$ with $$\hat{\epsilon}_{\theta, \phi}(x_t,t) = \epsilon_\theta(x_t,t) - s\sqrt{1-\overline{\alpha}_t}\cdot \nabla \log (\tau+ p_\phi(y=0|x_t, t)).$$  If $p_\phi(y=1|x_t,t) < \alpha$, replace $\epsilon_\theta$ with $$\hat{\epsilon}_{\theta, \phi}(x_t,t) = \epsilon_\theta(x_t,t) - s\sqrt{1-\overline{\alpha}_t}\cdot \nabla \log (\tau + p_\phi(y=1|x_t,t)).$$
    Otherwise, leave the sampling process unchanged.

By setting $\tau$ equal to 0.001, we were able to prevent the undesirable behavior.

\paragraph{Similarity Reduction Evaluation.} We employ the fine-tuned denoiser model to generate 3000 image samples for each of the aforementioned datasets. Additionally, we utilize the Classifier-guided method to generate another set of 3000 images. Subsequently, we employ DINO~\cite{caron2021emerging} to find nearest neighbors in the subset using a methodology akin to ambient diffusion. From the perspectives of both DINO's similarity scores and human evaluation, we observe that images generated through the classifier-guided approach exhibit significantly lower similarity to the original images in the subset compared to those generated without guidance.

\paragraph{FID Evaluation.} For each dataset, we utilize the denoiser model fine-tuned on the subset to generate 30,000 images under the guidance of the classifier. Subsequently, we employ the FID score implementation from the EDM~\cite{karras2022elucidating} paper to compute the FID score.

\paragraph{Inference Speed}
Although speed was not a goal of our method, we provide some context for how fast it is compared to standard diffusion.  We do our comparison using a batch size of $1$ to generate $10$ images with $50$ denoising steps.  CPSample with $\alpha = 0.5$ (i.e. computing gradients of the classifier at every step) had an average per-image generation time of $26.1\pm 0.029 s$.  By contrast, standard stable diffusion had an average generation time of $23.92 \pm 0.055 s$.  Therefore, when the classifier is small compared to the size of the diffusion model, the added time cost is insignificant.

\section{Membership Inference Attacks}\label{permutation_test}
\begin{algorithm}
    \caption{\label{inference_attack}  Test statistic for membership inference attack against diffusion models \cite{matsumoto2023membership}}
    \begin{algorithmic}
        \State \textbf{Input:} Target samples $x_1,...,x_m$, CPSample denoiser $\hat{\epsilon}_{\theta, \phi}$, noise schedule $\overline{\alpha}_t = \prod_{s=1}^t(1-\beta_s)$
        \State total\_error $~\leftarrow~$ 0
        \For{$x$ in $\{x_1,...,x_m\}$}
        \State total\_error $~\leftarrow~$ total\_error + $\lVert \epsilon - \hat{\epsilon}_{\theta, \phi}(\sqrt{\overline{\alpha}_t} x + \sqrt{1-\overline{\alpha}_t}\epsilon, t)\rVert^2$
        \EndFor
        \State mean\_error $~\leftarrow~$ total\_error$/m$.
    \end{algorithmic}
\end{algorithm}

In keeping with our goal of preventing membership inference attacks that are based on high similarity to a single member of the training set, we also perform a permutation test to ensure that we are not producing images that are anomalously close to the training data.  Explicitly, we test the null hypothesis: generating images from CPSample produces images that are no more similar to the training data than they are to arbitrary points drawn from the data distribution.  Our tests are performed in the same setting used in Section~\ref{similarity_reduction_subset}. 
Let $S = \{x_1,...,x_k\}$ be the data used for fine-tuning.  Let $T=\{x_1,...,x_k, x_{k+1},...,x_n\}$ be the entire training set.  Finally, let $P=\{\Tilde{x}_1,...,\Tilde{x}_m\}$ be samples from CPSample.  Then our permutation test is as follows:
\begin{enumerate}
    \item Sample $\Tilde{x}_1,...,\Tilde{x}_k$ from $P$ without replacement. For each $\Tilde{x}_i$, compute the quantity in \ref{cosine_similarity} where the nearest neighbor is chosen among $S$.  Let the similarity score of the most similar pair be $a$.
    \item Repeat the following process $\ell$ times: sample $S^{i}\subset T$ without replacement from $T$ so that $|S^{i}| = k$.  Sample $P^{i}$ without replacement from $P$ so that $|P^{i}| = k$.  Compute the most similar image in $S^{i}$ for each member of $P^{i}$.  Call the similarity of the most similar pair $a_i$.  
    \item  For a pre-specified level $\alpha$, reject the null hypothesis if $\frac{1}{\ell}\sum_{i=1}^\ell \mathbf{1}\{a_0> a_i\} > \alpha$.
\end{enumerate}
The results can be found in Table \ref{perm_test}.  Note that the test fails to reject on CIFAR-10 and LSUN Church, but succeeds on CelebA.  This is likely because we fine-tuned the CelebA model more extensively than the other two.
\begin{table}[htbp]
  \centering
  \caption{\label{perm_test}Reduction in cosine similarity between generated images and nearest neighbor in fine-tuning data.}
  \setlength{\tabcolsep}{4pt}
  \begin{threeparttable}
       \begin{tabular}{ccccccc}
    \toprule
         Dataset & FT Steps & $\alpha$ & Scale & DDIM& CPSample \\\midrule CIFAR-10 & 150k& 0.001 & 1  & 0.92 & 0.47  \\ 
         CelebA & 650k & 0.001 & 1000 &  0.99 &0.99\\ LSUN Church & 455k & 0.1 & 10  & 0.99 & 0.60 \\\bottomrule
    \end{tabular}%
    \begin{tablenotes}
        \item[1] $p$-values were computed using a log rank test for $H_0$: CPSample did not reduce the fraction of images with similarity score exceeding the threshold.
    \end{tablenotes}
  \end{threeparttable}
\end{table}

\section{Additional Empirical Results}\label{empirical_dump}

\begin{table}[htbp]
  \centering
  \caption{FID score \emph{w.r.t.} $\alpha$ and $\text{Scale}$ on CIFAR-10.}
    \begin{tabular}{l|ccccc}
    \toprule
          & $\alpha=0.001$ & $\alpha=0.01$  & $\alpha=0.1$   & $\alpha=0.25$  & $\alpha=0.49$ \\
    \midrule
    $\text{Scale}=1$     & 4.14275 & 4.15467 & 4.19058 & 4.19208 & 4.21859 \\
    $\text{Scale}=5$     & 4.15772 & 4.20731 & 4.36005 & 4.58839 & 4.9566 \\
    $\text{Scale}=10$    & 4.18083 & 4.26594 & 5.05858 & 6.17326 & 7.88949 \\
    $\text{Scale}=100$   & 4.96727 & 16.7173 & 74.7247 & 113.199 & 139.626 \\
    \bottomrule
    \end{tabular}%
  \label{fid_effects}%
\end{table}%

\begin{figure}[H]
    \centering
    \includegraphics[width=0.6\linewidth]{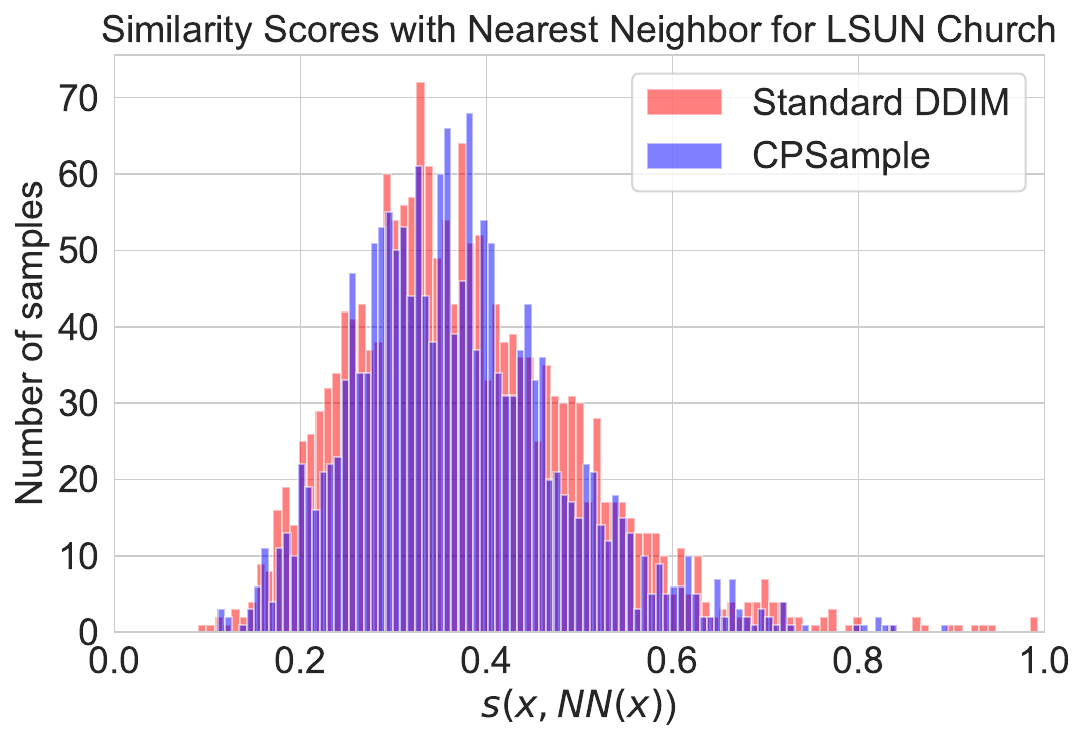}
    \caption{Similarity scores with nearest neighbor for standard DDIM and CPSample ($\alpha=0.1$, scale$=10$) on LSUN Church.  In both cases, the network was fine-tuned for 455k gradient steps on a subset of $1000$ images. }
    \label{fig:lsun_church_reduction}
\end{figure}

\begin{figure}[H]
    \centering
    \includegraphics[width=1\linewidth]{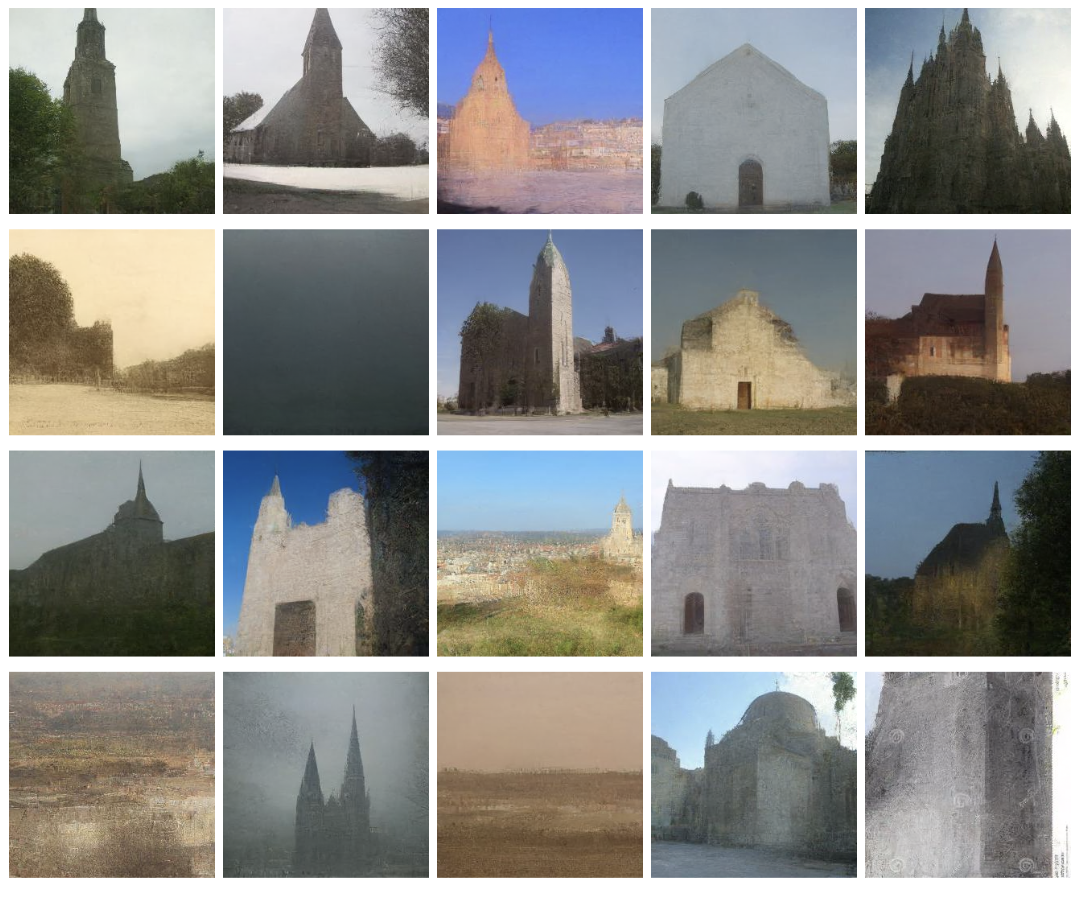}
    \caption{Uncurated samples using standard DDIM fine-tuned for 455k gradient steps on a subset of 1000 images from LSUN Church.  }
    
\end{figure}

\begin{figure}[H]
    \centering
    \includegraphics[width=1\linewidth]{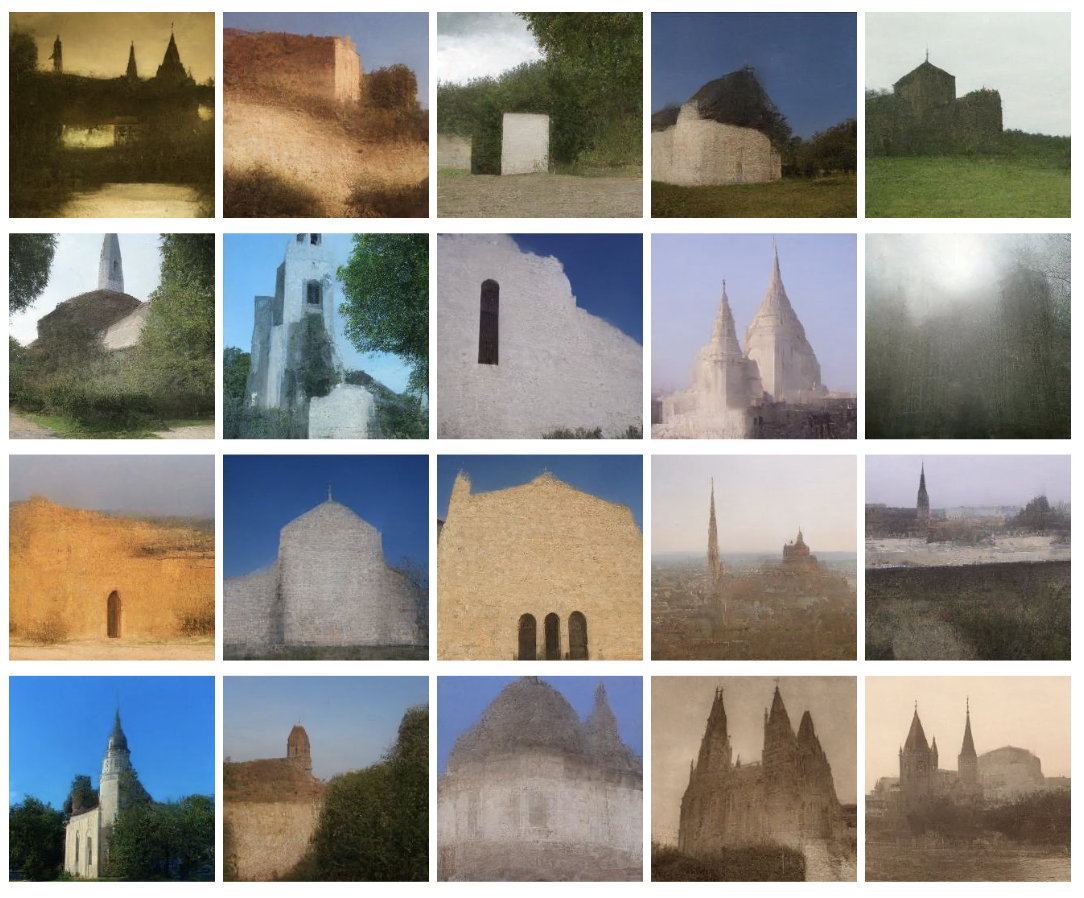}
    \caption{Uncurated samples using CPSample ($\alpha = 0.1$, scale$=10$) applied to a network fine-tuned for 455k gradient steps on a subset of 1000 images from LSUN Church. Note that there is no visual discrepancy in quality between these and the images from standard DDIM. }
    
\end{figure}

\begin{figure}[H]
    \centering
    \includegraphics[width=1\linewidth]{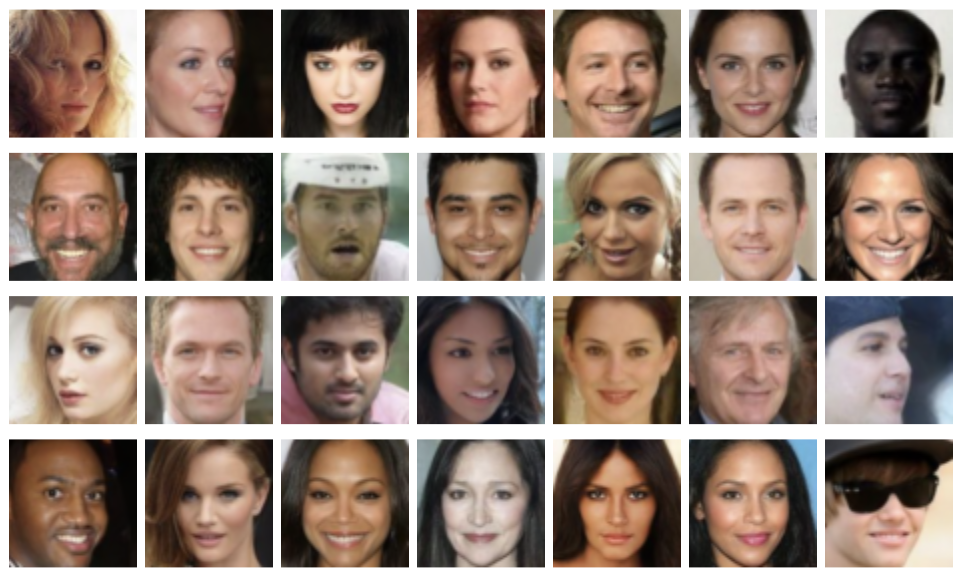}
    \caption{Uncurated samples using standard DDIM fine-tuned for 580k gradient steps on a subset of 1000 images from CelebA. }
    
\end{figure}

\begin{figure}[H]
    \centering
    \includegraphics[width=1\linewidth]{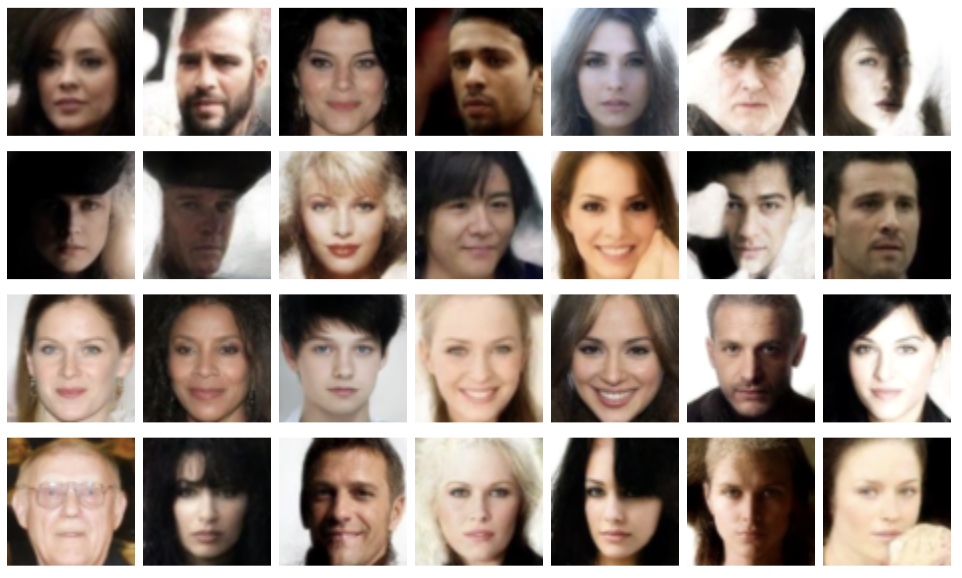}
    \caption{Uncurated samples using CPSample ($\alpha = 0.001$, scale$=1000$) applied to a network fine-tuned for 580k gradient steps on a subset of 1000 images from CelebA. Note that there is little visual discrepancy in quality between these and the images from standard DDIM. }
    
\end{figure}

\begin{figure}[H]
    \centering
    \includegraphics[width=1\linewidth]{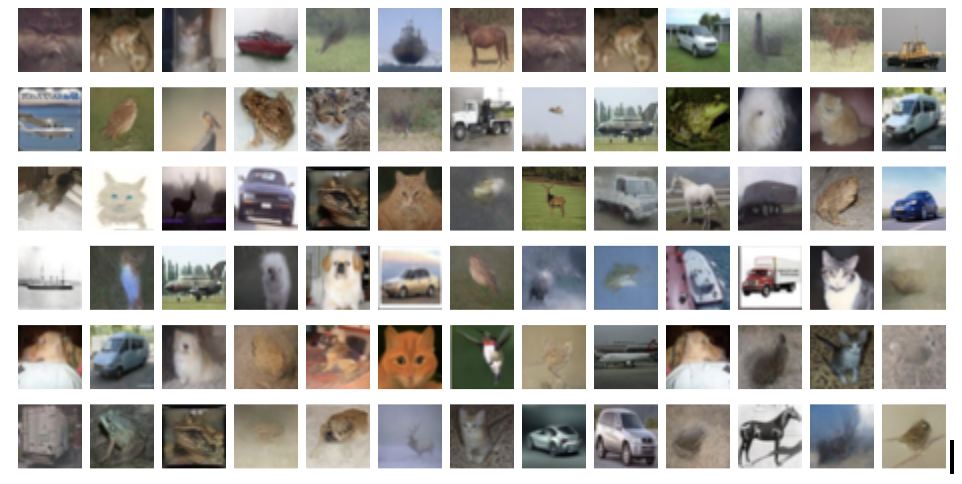}
    \caption{Uncurated samples using standard DDIM fine-tuned for 150k gradient steps on a subset of 1000 images from CIFAR-10. }
    
\end{figure}

\begin{figure}[H]
    \centering
    \includegraphics[width=1\linewidth]{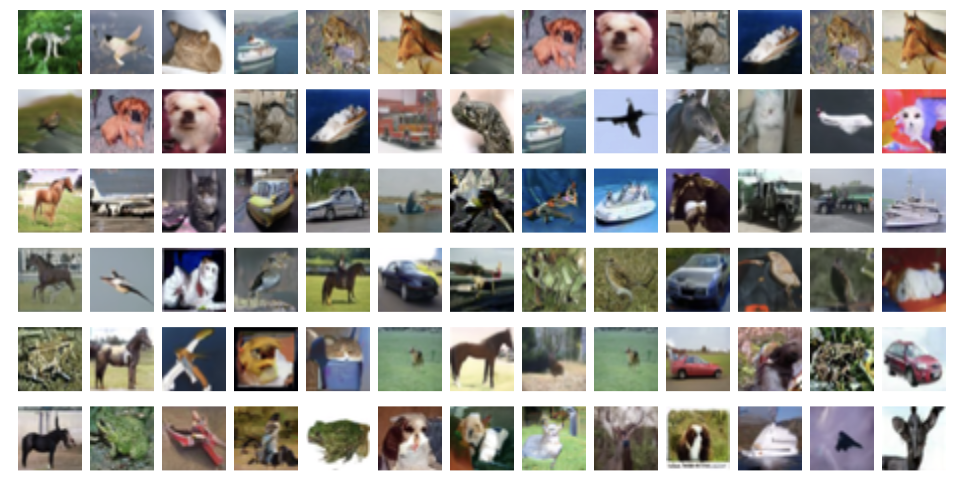}
    \caption{Uncurated samples using CPSample ($\alpha = 0.001$, scale$=1$) applied to a network fine-tuned for approximately 150k gradient steps on a subset of 1000 images from CelebA. Note that there is little visual discrepancy in quality between these and the images from standard DDIM. }
    
\end{figure}

\end{document}